\pdfoutput=1
\documentclass[11pt]{article}

\usepackage{acl}

\usepackage{times}
\usepackage{latexsym}

\usepackage[T1]{fontenc}
\usepackage[utf8]{inputenc}

\usepackage{microtype}

\usepackage{graphicx}

\usepackage{amssymb, amsmath}

\usepackage{pifont}
\newcommand{\cmark}{\ding{51}}%
\newcommand{\xmark}{\ding{55}}%

\usepackage{tabularx}

\usepackage{subcaption}

\usepackage{booktabs}

\usepackage{needspace}
\usepackage{float}
\usepackage{amsmath}
\usepackage{placeins}

\title{Racial Bias Trends in the Text of US Legal Opinions}

\author{Rohan Jinturkar \\
    Princeton University \\ \texttt{rohanj@princeton.edu}}
    
\begin{document}
\maketitle
\begin{abstract}
Although there is widespread recognition of racial bias in US law, it is unclear how such bias appears in the language of law, namely judicial opinions, and whether it varies across time period or region. Building upon approaches for measuring implicit racial bias in large-scale corpora, we approximate GloVe word embeddings for over 6 million US federal and state court cases from 1860 to 2009. We find strong evidence of racial bias across nearly all regions and time periods, as traditionally Black names are more closely associated with pre-classified ``unpleasant'' terms whereas traditionally White names are more closely associated with pre-classified ``pleasant'' terms. We also test whether legal opinions before 1950 exhibit more implicit racial bias than those after 1950, as well as whether opinions from Southern states exhibit less change in racial bias than those from Northeastern states. We do not find evidence of elevated bias in legal opinions before 1950, or evidence that legal opinions from Northeastern states show greater change in racial bias over time compared to Southern states. These results motivate further research into institutionalized racial bias.
\end{abstract}

\section{Introduction and Motivation}
Historically, US legal institutions have upheld racial inequality. Many past judicial decisions, including \emph{Dred Scott v. Sandford} (descendants of slaves are not citizens)\footnote{https://www.oyez.org/cases/1850-1900/60us393}, \emph{Plessy v. Ferguson} (state-imposed segregation is legal)\footnote{https://www.oyez.org/cases/1850-1900/163us537}, and \emph{Korematsu v. United States} (wartime internment of Japanese-Americans is legal)\footnote{https://www.oyez.org/cases/1940-1955/323us214}, are today considered racially discriminatory. Despite advances in civil rights, Americans remain skeptical about the state of racial equality: in a 2019 Pew Research Poll, 87\% of Black respondents and 61\% of White respondents believed Black individuals were treated less fairly by the justice system. 84\% of Black respondents and 63\% of White respondents believed that Black individuals were treated less fairly by the police \cite{pew}.

Given the US legal system's historical influence on racial justice efforts, it is important to analyze a potential medium of implicit racial bias: the text of judges' opinions, which explain a court's decision. Because law is shaped by the precise language of judicial opinions, examining their phrasing and meaning is crucial \citep{legaldoctrine}. Legal opinions also contain language that often convey judges' emotions \citep{rice_zorn_2021}. 

The presence of racial bias in legal text has serious implications. First, evidence of bias raises questions about the fairness of the legal proceedings that led to the decision \citep{legalbias}. Second, because judicial opinions create precedents by shaping existing legal principles, there are downstream effects as legislators, judges, police, and other players adjust their behavior \citep{hansford_spriggs_2006}.

To the best of our knowledge, there is no prior research examining racial bias trends in the text of US legal opinions across region or time period. By analyzing opinions from 1860 to 2009, segmented by five regions (Northeast, West, South, Midwest, Federal) and six equally sized time intervals (1860-1889, 1890-1919, 1920-1949, 1950-1979, 1980-2009), we can improve our understanding of textual bias in the US legal system. This study explores three research questions. First, do US legal opinions exhibit implicit racial bias across all regions and time periods? Second, do older legal opinions (e.g., before 1950) exhibit greater racial bias than newer legal opinions? Third, do opinions from Southern states (which historically exhibited greater post-Civil War resistance to racial equality) exhibit less change in racial bias over time than those from Northeastern states?

We offer two contributions. First, we present an approach to compare relative bias between corpora with different data distributions. Second, we support analysis of whether the content of decisions and deliberations reflects shifts in societal attitudes across time and region. For instance, do judges' opinions today reflect a more positive view of people of color than they did in the 1860s? The experiments conducted in this study can help determine whether progress toward racial equality is evenly distributed across the United States.

\section{Related Work}
% \textbf{general methods for identifying bias}
% Many methods exist to quantitatively assess racial bias. \citet{goff_police} found a statistically significant disparity in the use of force between arrests of White and Black individuals. \citet{yang_sentencing} uncovered racial disparities in sentence lengths following \emph{United States v. Booker} (2005), which increased judicial discretion in sentencing. Yang found that following the ruling, Black defendants received two more months in prison on average compared to White defendants.

%Notably, mean use of force rates per 100,000 residents (when controlled for race) were approximately three times as high for Black individuals as for White individuals. This discrepancy held true even when controlling for arrest demographics, providing strong evidence of racial disparities in the use of force during arrests. - NOTE: NOT NLP RELATED?

%, with the disparity primarily driven by a greater reliance on mandatory minimum sentences against Black defendants. - NOT NLP RELATED?

To identify implicit forms of bias, \citet{Greenwald1998MeasuringID} introduced the Implicit Association Test (IAT) to measure the association between concepts (e.g., race, age, gender) and evaluations (e.g., good, bad). When applied to the legal system, the IAT indicated that participants consistently held strong implicit associations between Black men and guilty verdicts \citep{levinson}. 

Identifying a concept like race or age in text can be difficult when they are not explicitly mentioned, such as in legal documents. A popular approach in cognitive and political science studies is to use racially distinctive names to signal the individual's race (\citealp{bertrand_mullainathan_2004}; \citealp{merullo-etal-2019-investigating}). \citet{butler_homola_2017} empirically showed that an individual's response to a racially distinctive set of names is driven by the implied race and not other factors, justifying their use as a proxy in a corpus.  

Along with the concepts, there is a need for a target set of pre-classified pleasant and unpleasant terms (the ``evaluation'') to use in IAT studies. The AFINN dataset contains 24777 English words and phrases with a sentiment polarity score \citep{AFINN}, where positive terms occur in a pleasant context and negative terms are associated to less pleasant contexts. 

Some studies have utilized racially distinctive names and emotion lexicons like the AFINN dataset to identify bias in corpora through GloVe embeddings, i.e., a closer proximity between a set of European American names to ``pleasant'' terms than a set of African American names (\citealp{embeddings_bias}; \citealp{friedman-etal-2019-relating}). \citet{embeddings_bias} also proposed the Word Embedding Association Test (WEAT): using the null hypothesis that there is no difference in the mean similarity between two sets of words (e.g., STEM vs. non-STEM) and two sets of attributes (e.g., men vs. women), the test yielded a test statistic through which the hypothesis could be evaluated. \citet{lauscher-glavas-2019-consistently} compared WEATs across multiple embeddings, finding that GloVe embeddings can accentuate biases while others, like \texttt{FastText}, diminish them. As such, GloVe has become the standard for bias analysis. 

Because the true distribution of results may be unknown when the null hypothesis is true (e.g., true distance between spurious associations), \citet{Lax2010LegalCO} proposed the \textit{randomization test} to generate a reference distribution. Hypothesis tests typically make explicit assumptions regarding the shape of a distribution and correlations between variables, but randomization tests do not. The authors shuffled the data to break any correlation between the treatment and outcome, and calculated a test statistic. By repeating this process many times, they generated a distribution of test statistics that would be observed if the null hypothesis was true, i.e., no relationship between the treatment and outcome. If the true test statistic exceeded some threshold $\alpha$, the result was considered statistically significant.

\citet{legalbias} utilized the above approaches on a corpus of US federal and state opinions, training GloVe embeddings and assembling a set of traditionally Black/White names and pleasant/unpleasant terms. They performed a WEAT to capture relative associations of names to terms. Since the distribution of associations was unknown under the null hypothesis, they performed randomization tests using the approach proposed by \citet{Lax2010LegalCO}. They found that African American names were more frequently associated with pre-classified unpleasant concepts whereas White names were more frequently associated with pre-classified pleasant concepts, offering evidence of implicit racial bias.

The authors did not examine whether such bias varies by time period or region (this was mentioned as a goal for future work). We build on this research to conduct a historical analysis and inform future bias mitigation efforts.

\section{Approach}

\subsection{Data Collection}
Opinions are sourced from the Harvard Law School Caselaw Access Project \citep{caselaw}. It contains a digitized collection of all official, book-published state and federal US court proceedings. It does not include records of lower court opinions, so they are excluded from this study. 

We group the cases by time (30 year intervals from 1860-2009) and region (US Census Bureau classifications). The dataset contains approximately 6.2 million cases, each with one judicial opinion. State-level cases are clustered by one of four US census regions, and federal-level cases are treated as their own region. Since there are five regions and six time periods, there are 30 unique corpora in total. The case count breakdown by time period and region is provided in Appendix D. 

\begin{figure}[hbt]
\centering
\includegraphics[width=0.85\linewidth]{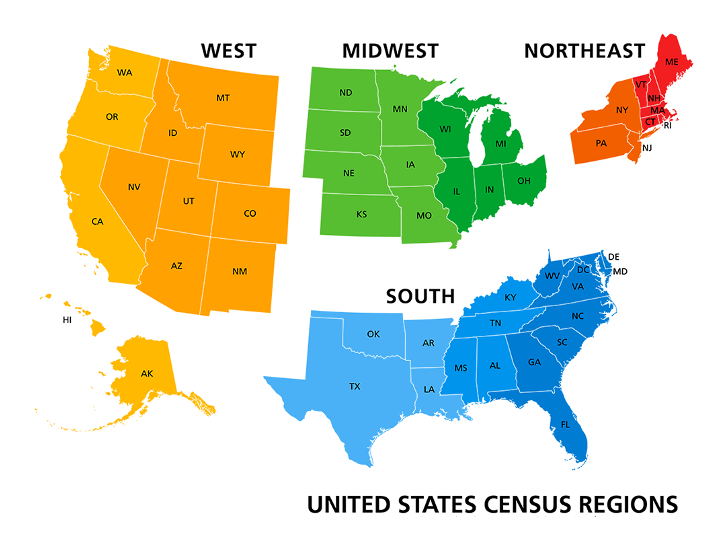}
\caption{Region splits \citep{censusregions}.}
\label{fig:regions}
\end{figure}

We map each state to its corresponding US Census Bureau region in Figure \ref{fig:regions}. Then we query the Caselaw Access Project API endpoint for all cases for states in a given region within the specified time period. After standard preprocessing (described in Appendix A), we store the cases in individual text files. 

\subsection{Word Embeddings}
We leverage GloVe embeddings \citep{pennington-etal-2014-glove}, which have been used for similar studies into legal bias. For each corpus, we exclude words that occur fewer than 10 times, set the co-occurrence window to 20 tokens, and train 200-dimensional vectors for a maximum of 15 iterations, following \citet{legalbias}. The number of unique embeddings per corpus is shown in Table \ref{table:word_counts}. 

\begin{table*}[hbt]
  \centering
  \caption{Vocabulary size as the number of words occurring at least 10 times per corpus.}
  \label{table:word_counts}
  \renewcommand{\arraystretch}{1.2}
  \begin{tabular}{rrrrrr}
    \toprule
      & \textbf{Northeast} & \textbf{South} & \textbf{Midwest} & \textbf{West} & \textbf{Federal}\\
    \midrule 
    \textbf{1860 - 1889} & 42605 & 44957 & 45524 & 18929 & 30090\\
    \textbf{1890 - 1919} & 76461 & 86031 & 91861 & 48895 & 56790\\
    \textbf{1920 - 1949} & 71528 & 116491 & 94689 & 67825 & 91601 \\
    \textbf{1950 - 1979} & 100166 & 131959 & 113607 & 96726 & 142543\\
    \textbf{1980 - 2009} & 221376 & 318293 & 241266 & 176720 & 347592\\
    \bottomrule
  \end{tabular}
  % \caption*{NE = Northeast, S = South, MW = Midwest, W = West, F = Federal}
  \vspace{-4mm}
\end{table*}
\subsection{Indicators of Race}
A key challenge is to identify the race of individuals discussed in court documents, since opinions rarely include explicit mentions of race. A potential signal is the usage of African American English (AAE). However, AAE does not work here because opinions are typically written by the judge and rarely include quoted statements from a plaintiff or defendant. On the other hand, legal opinions explicitly include the names of plaintiffs and defendants, so those can be used as a signal for race. This choice follows the bias-testing framework used by \citet{embeddings_bias} and \citet{legalbias}. 

Most names (e.g., Jasmine, David) are noisy signals for race, i.e., there may not be a strong correlation between the name and a particular race. Conversely, a racially distinctive set of names will not represent the true distribution of names in the population. Given the importance of understanding racial bias in legal contexts and the absence of stronger signals, we use the names of individuals mentioned in court proceedings.  We provide more discussion in the Limitations section.

We use the set of names proposed by \citet{butler_homola_2017} and tested by \citet{legalbias}. Since these names were used for a racial bias study of legal text, we assume the same set is a good basis for our study as well. The complete list of names is provided in Appendix B. The total number of names used is relatively small to ensure they are sufficiently strong indicators for race. The number of names identified from our set in each corpus is listed in Table \ref{table:individual_counts}.

\begin{table}[hbt!]
    \centering
  \caption{Number of distinctive names by corpus with at least 10 occurrences.}
  \label{table:individual_counts}
  \renewcommand{\arraystretch}{1.2}
  \begin{tabular}{rrrrrr}
    \toprule
      & \textbf{NE} & \textbf{S} & \textbf{MW} & \textbf{W} & \textbf{F}\\
    \midrule 
    \textbf{1860 - 1889} & 9 & 9 & 10 & 5 & 7\\
    \textbf{1890 - 1919} & 13 & 12 & 12 & 9 & 9\\
    \textbf{1920 - 1949} & 12 & 14 & 12 & 11 & 10 \\
    \textbf{1950 - 1979} & 16 & 18 & 16 & 17 & 16\\
    \textbf{1980 - 2009} & 31 & 32 & 31 & 27 & 28\\
    \bottomrule
     \end{tabular}
     \caption*{NE = Northeast, S = South, MW = Midwest, W = West, F = Federal.}
  \vspace{-4mm}
\end{table}

\subsection{Sentiment Terms}
Next, we compile a set of pleasant and unpleasant terms for which to measure the distance from Black and White names. We utilize the AFINN dataset, using each term's given sentiment polarity to map it as either pleasant or unpleasant \cite{AFINN}.

The AFINN dataset includes a few terms with strong legal connotations, such as ``murder'' and ``freedom.'' These terms were labeled in a non-legal context, so their valence might be different within a legal setting. This would be a problem if the dictionary was small, as terms that do not hold the same pleasant/unpleasant meaning would decrease the likelihood of identifying the effect. However, because the AFINN dataset contains over 2400 unique terms/phrases, the relatively large size of the dictionary outweighs the negative influence of the few legal terms present. 

\subsection{Method}
We test the null hypothesis (via the IAT) that there is no relationship between Black/White names and their mean distance to pleasant/unpleasant terms. Because we do not know the distribution of associations between the names and terms if there is no relationship, we use randomization tests (explained further below) to determine statistical significance.  

First, we estimate the association between each traditionally Black or White name and each pleasant or unpleasant term. We follow the approach taken by \citet{embeddings_bias} and \citet{legalbias} to calculate a word association measure.

\begin{equation}
\label{eq:word_assoc}
    s(w, A, B) = \frac{1}{|A|} \sum_{a \in A} cos(\vec{w}, \vec{a}) - \frac{1}{|B|} \sum_{b \in B} cos(\vec{w}, \vec{b})
\end{equation}

\begin{equation}
    \label{eq:cos_sim}
    cos(\vec{a}, \vec{b}) = \frac{\vec{a} \cdot \vec{b}}{\| \vec{a} \|  \|  \vec{b} \|}
\end{equation}

For each name $w$, set of pleasant words $A$, and set of unpleasant words $B$, we calculate the word association $s(w, \ A, \ B)$ as shown in Equation \ref{eq:word_assoc} ($\vec{w}$ represents the word embedding vector for $w$). Equation \ref{eq:cos_sim} shows the cosine similarity calculation. If a name is more closely associated with pre-classified pleasant words than unpleasant words, $s(w, A, B)$ will be positive. Per our pre-processing, names or categories that occur fewer than 10 times in the corpus do not have a vector embedding, so they are skipped.

The above calculation yields a measure of the relative association of names to categories. However, this measure lacks context; what should the distribution of associations between the sets of names and terms look like if there is no relationship between them? To solve this, we utilize the randomization tests proposed by \citet{Lax2010LegalCO}: we generate a distribution of test statistics that would exist if the null hypothesis (no relationship) is true, and then plot the actual test statistic against it to determine if the value is statistically significant.

Similar to \citet{embeddings_bias} and \citet{legalbias}, we first calculate a test statistic as depicted by Equation \ref{eq:test_stat}. $X$ and $Y$ represent the set of Black and White name vectors, respectively.

\begin{equation}
\label{eq:test_stat}
    s(X,Y,A,B) = \sum_{x \in X} s(x,A,B) - \sum_{y \in Y} s(y,A,B)
\end{equation}

If the set of traditionally White names co-occurs more with pleasant terms and the set of traditionally Black names co-occurs more with unpleasant terms, then the test statistic will be a large positive value. 

We next take the combined set of pleasant/unpleasant words and randomly assign them to pleasant and unpleasant categories. In other words, we throw out the existing label to evaluate a test statistic asking whether there was no relationship between the names and their mean word association to a given category. We repeat the above process 1000 times, shuffling the words in each category and calculating a new test statistic. The distribution of randomization test statistics is approximately normal.

If the originally observed test statistic (using the true labels for pleasant and unpleasant words) is at the right tail of this \emph{randomized} distribution (threshold of $\alpha = 0.05$), the result is considered statistically significant. Such a result would mean that commonly Black names co-occur closer to unpleasant terms whereas commonly White names co-occur closer to pleasant terms and offer evidence of implicit racial bias against Black individuals. 

\subsubsection{Comparison Across Subsets} The above experiments test for racial bias within corpora but not between them. Because GloVe embeddings are unique to each corpus, the test statistics cannot be directly compared against each other. Instead, we generate a modified test statistic reflecting the normalized difference across corpora $V$ and $W$, shown in Equation \ref{eq:stat_diff}:

\begin{equation}
    \label{eq:stat_diff}
    T(V, W) = \frac{T^{V}_{obs}}{\sqrt{var(T^{V}_{random})}} - \frac{T^{W}_{obs}}{\sqrt{var(T^{W}_{random})}}
\end{equation}

$T^{V}_{obs}$ and $T^{W}_{obs}$ are the observed test statistics for corpora $V$ and $W$, and the denominator of both terms is the standard deviation of the distribution generated by the randomization tests from earlier. To contextualize the difference, we modify the randomization tests as proposed by \citet{congressbias} in a study of Congressional speeches.
 
First, we randomly generate 20 new corpora, ranging in size from 1 to 2 gigabytes, by combining random segments from existing corpora. For each new corpus, we generate a word embedding test statistic as shown in Equation \ref{eq:test_stat}. Applying Equation \ref{eq:stat_diff}, we estimate the difference across all combinations, yielding $\binom{20}{2} = 190$ differences.

The remaining step is to compute the difference in association between two corpora (e.g., two regions from 1980-2009) and plot the value on the distribution. If this value is greater than  95\% of the 190 data points from the randomization tests, it is considered statistically significant.

\section{Results and Discussion}
\subsection{Term Frequency}
Cross-sections of the most common words across time periods and regions are provided in Tables \ref{table:common_time} and \ref{table:common_region}. The following terms were in the top 10 across all 30 corpora: court, case, defendant, plaintiff, law. 

\begin{table}[hbt!]
    \centering
    \setlength{\tabcolsep}{1.3em}
    \setlength{\extrarowheight}{0.3em}
    \caption{Most common terms appearing across each time period (frequency order).}
    \begin{tabularx}{\linewidth}{lX}
        \toprule
            \textbf{\small{Time Period}} & \textbf{\small{Terms}} \\
        \midrule
            \small{1860-1889} & \small{court, case, defendant, plaintiff, judgment, law, state, act, right, property} \\
        \midrule
            \small{1890-1919} & \small{court, defendant, plaintiff, case, state, law, evidence, judgment, time, fact} \\
            
        \midrule
            \small{1920-1949} & \small{court, defendant, plaintiff, case, state, time, evidence, appellant, judgment, law} \\
        \midrule
            \small{1950-1979} & \small{case, court, defendant, evidence, fact, law, order, plaintiff, state, time} \\
        \midrule
            \small{1980-2009} & \small{court, defendant, state, trial, case, plaintiff, evidence, claim, order, law} \\
        \bottomrule
    \end{tabularx}
    \label{table:common_time}
\end{table}
    
\begin{table}[hbt!]
    \centering
    \setlength{\tabcolsep}{1.3em}
    \setlength{\extrarowheight}{0.3em}
    \caption{Most common terms appearing across each region (frequency order).}
    \begin{tabularx}{\linewidth}{lX}
        \toprule 
            \small{\textbf{Region}} & \small{\textbf{Terms}} \\
        \midrule
            \small{Northeast} & \small{court, defendant, plaintiff, case, state, law, order, trial, evidence, judgment} \\
        \midrule
            \small{South} & \small{court, state, defendant, trial, case, evidence, appellant, plaintiff, judgment, law} \\
        \midrule
            \small{Midwest} & \small{court, defendant, state, plaintiff, case, trial, evidence, judgment, law, time} \\
        \midrule
            \small{West} & \small{court, defendant, state, case, trial, plaintiff, evidence, judgment, law, time} \\
        \midrule
            \small{Federal} & \small{court, plaintiff, defendant, case, state, claim, united, law, act, district} \\
        \bottomrule
    \end{tabularx}
    \label{table:common_region}
\end{table}

The most common terms barely change across time periods and regions, suggesting that the US judicial lexicon is very stable. 

\subsection{Bias within each Corpus}
Our first research question is whether US legal opinions exhibit implicit racial bias across all regions and time periods. Table \ref{table:bias regions} depicts the results of randomization tests (i.e., if there is no relationship between the race by name and mean association to pre-classified pleasant/unpleasant terms).

\renewcommand{\arraystretch}{1.1}% for the vertical padding
\begin{table}[hbt]
  \centering
  \caption{Summary of randomization test results for bias within a corpus.}
  \begin{tabular}{rr@{\hskip 0.2in}r@{\hskip 0.15in}r@{\hskip 0.2in}r@{\hskip 0.2in}r@{\hskip 0.2in}}
    \toprule
       & \textbf{NE} & \textbf{S} & \textbf{MW} & \textbf{W} & \textbf{F}\\
    \midrule 
    \textbf{1860 - 1889} & \cmark & \cmark & \cmark & \xmark & \xmark\\
    \textbf{1890 - 1919} & \cmark & \xmark & \xmark & \cmark & \cmark\\
    \textbf{1920 - 1949} & \cmark & \cmark & \cmark & \cmark & \cmark\\
    \textbf{1950 - 1979} & \cmark & \cmark & \cmark & \cmark & \cmark\\
    \textbf{1980 - 2009} & \cmark & \cmark & \cmark & \cmark & \cmark\\
    \bottomrule
  \end{tabular}
  \caption*{\cmark = statistically significant, \xmark = not statistically significant for randomization test, $\alpha \leq 0.05$. NE = Northeast, S = South, MW = Midwest, W = West, F = Federal.}
  \label{table:bias regions}
  \vspace{-6mm}
\end{table}

In almost every corpus, the result is statistically significant (the WEAT distribution charts are provided in Appendix F). The test statistics fall on the right end of the generated distribution, indicating that traditionally White names more frequently co-occur with pleasant terms and traditionally Black names more frequently co-occur with unpleasant terms. These results indicate the presence of implicit racial bias across region/time slices.  

There are four corpora for which the null hypothesis is not rejected: the West and Federal regions from 1860-1889, and the South and Midwest regions from 1890-1919.

One explanation is that there are fewer available opinions for older time periods. Per the case counts table (in Appendix D), West (1860-1889) had 17560 cases, Federal (1860-1889) had 28123 cases, South (1890-1919) had 190129 cases, and Midwest (1890-1919) had 191487 cases. The former two corpora are relatively small, which could explain why fewer names were found in the text. As Table \ref{table:individual_counts} shows, the newer corpora almost consistently have more unique names with at least 10 occurrences in the corpus. For corpora with fewer names, we would expect the test statistic to be close to zero, which gives inconclusive evidence of racial bias.

Another explanation is that the composition of names has changed over time. We use a contemporary set of Black and White names to measure bias across all corpora, but names change in popularity. Consequently, a larger proportion of the names may not have been present in the corpora of older legal opinions, reducing the number of signals for race and providing less information about implicit racial bias. We track the semantic changes of the names used.

The Google Books Ngram Viewer\footnote{https://books.google.com/ngrams} provides a rough estimate of the changing popularity of terms over time \citep{google_books}. This metric is slightly noisy: we expect that commonly Black names would be more recently popular in text because there are historically very few published works by Black authors and/or centered around Black communities. However, it can still indicate whether the changing popularity of names should be incorporated in future work. We assume that the popularity of names is reflected by their frequency of occurrence in text, i.e., more popular names will appear more frequently in the Google Ngram Viewer. 

We find that some of our names, such as ``Aaliyah,'' only recently became popular and is also not in the vocabulary for any region from 1860-1889. The Google Ngram Viewer output for these names can be found in Appendix E. The changing popularity of names would also correlate with our observation from Table \ref{table:individual_counts}: the older corpora have fewer unique names identified, while the newer corpora have more unique names. 

\subsection{Relative Bias Across Corpora}
Below are the results demonstrating how racial bias has changed across region and time period. 

\begin{figure}[hbt]
\centering
\includegraphics[width=\linewidth]{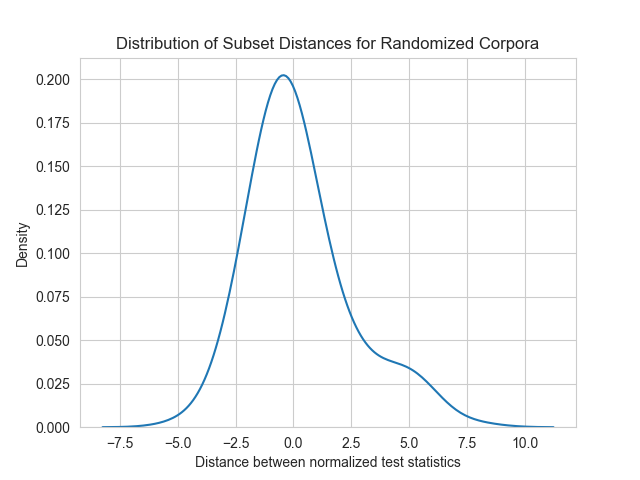}
\caption{Distribution of differences in test statistic distances for randomized corpora.}
\label{fig:diff_randomized}
\end{figure}
%\textbf{Add average distances for selected pairs as dots? E.g., interesting outliers?}

Figure \ref{fig:diff_randomized} shows the distribution of subset distances between all combinations of the randomly generated corpora. The full list of test statistics, as calculated by Equation \ref{eq:stat_diff}, can be found in Appendix C. To determine if the relative bias between two corpora (calculated from Table \ref{table:bias regions}) is statistically significant, we plot the difference on the distribution in Figure \ref{fig:diff_randomized}.

\renewcommand{\arraystretch}{1.2}
\begin{table*}[hbt]
  \centering
  %\setlength{\tabcolsep}{1.3em} % for the horizontal padding
  %\setlength{\extrarowheight}{0.8em}
  %{\renewcommand{\arraystretch}{1.2}% for the vertical padding
  \caption{Summary of normalized test statistics for word embedding association test.}
  \label{table:normalized_stats}
  \begin{tabular}{rrrrrr}
    \toprule
      & \textbf{Northeast} & \textbf{South} & \textbf{Midwest} & \textbf{West} & \textbf{Federal}\\
    \midrule 
    \textbf{1860 - 1889} & 3.6170 & 2.8915 & 2.7966 & -0.7107 & 0.8754\\
    \textbf{1890 - 1919} & 4.4482 & 1.2455 & 0.6241 & 4.0155 & 4.5317\\
    \textbf{1920 - 1949} & 2.9944 & 1.8207 & 2.9423 & 4.1812 & 4.0246\\
    \textbf{1950 - 1979} & 2.0107 & 1.6749 & 4.2747 & 2.6782 & 3.8304\\
    \textbf{1980 - 2009} & 4.4286 & 5.3689 & 4.7148 & 3.1796 & 3.9874\\
    \bottomrule
  \end{tabular}
  %\caption*{NE = Northeast, S = South, MW = Midwest, W = West, F = Federal}
  \vspace{-4mm}
\end{table*}

\subsubsection{Older vs. Newer Corpora Comparison}
Our second research question is whether older legal opinions (pre-1950) exhibit more racial bias than newer legal opinions. We evaluate it by comparing the normalized test statistics calculated via Equation \ref{eq:stat_diff} across time periods.

\begin{figure}[hbt]
    \centering
    \includegraphics[width=\linewidth]{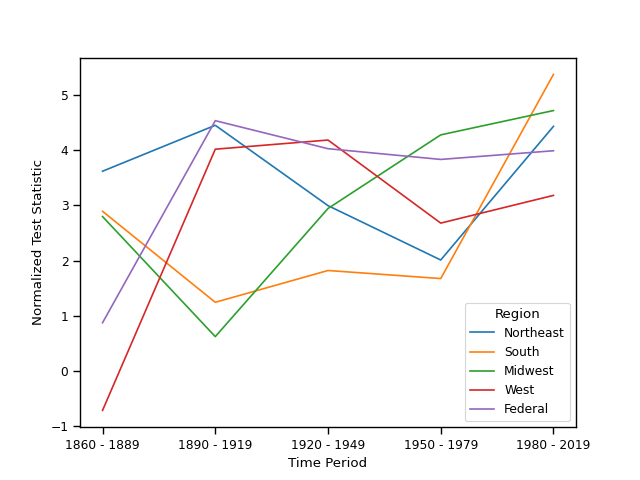}
    \caption{Normalized test statistic by region, 1860-2009.}
    \label{fig:normalized_stats}
\end{figure}

Table \ref{table:normalized_stats} and Figure \ref{fig:normalized_stats} paint a complicated picture. The test statistics for the oldest legal opinions do not uniformly point to racial bias; as discussed previously, this is likely due to the changing popularity of the names used. As such, our results are more likely to be valid for newer corpora, so we examine those immediately preceding and following 1950. 

For each region, we compare the normalized test statistic from 1920-1949 with that from 1950-1979 in Table \ref{table:normalized_stats}. We find that the latter corpus has less racial bias in all regions except the Midwest, where the test statistic increases. However, this trend reverses for the most recent corpora, where all regions show greater racial bias in 1980-2009 compared to 1920-1949. Since there is no consistent upward or downward trend, performing randomization tests will not yield additional information; there is inconclusive evidence that older opinions exhibit more racial bias than newer opinions. 

\subsubsection{Northeast vs. South Comparison}
Our third research question is whether opinions from Southern states exhibit less change in racial bias compared to those from Northeastern states. 

\begin{figure}[hbt]
\centering
\includegraphics[width=\linewidth]{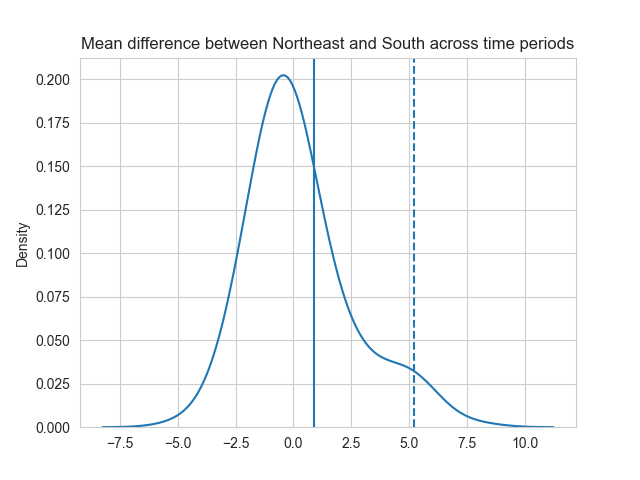}
\caption{Significance of mean test statistic difference between Northeast and South, all time periods.}
\label{fig:diff_ne_south}
\end{figure}

Based on a review of Table \ref{table:normalized_stats}, the Northeast almost consistently has a higher test statistic than the South. To determine if this difference is statistically significant, we compute the average of the pairwise distances between the normalized test statistics (from Equation \ref{eq:stat_diff}) for the Northeast and South across each of the six time periods and plot it on the distribution from Figure \ref{fig:diff_randomized}. We also plot the 5\% significance threshold (approximately 10 more extreme data points, out of 190 total) in Figure \ref{fig:diff_ne_south}. The mean difference is the solid line whereas the threshold is the dotted line.

The mean difference in the test statistic is not statistically significant, indicating that the level of implicit racial bias is not meaningfully different between the Northeast and South across time periods. 

Lastly, we determine if the change in racial bias for Southern opinions is less than that for Northern opinions. Both the Northeast and South seem to exhibit an \emph{increase} in racial bias over time, which can be attributed in part to the use of names that only recently became popular (discussed previously). This may also explain why the normalized test statistics are higher (i.e., stronger evidence of racial bias) in the most recent corpora for both the Northeast and South.

The relative difference in racial bias can be computed using Equation \ref{eq:stat_diff} with the newest and oldest corpora for the South and Northeast, respectively.

\begin{equation*}
    \begin{split}
    T_{South}(1980-2009, 1860-1889) = 2.4774 \\
    T_{Northeast}(1980-2009, 1860-1889) = 0.8116
    \end{split}
\end{equation*}

The change in racial bias in the South (2.4774) is greater than that of the North (0.8116). This might be influenced by the region clustering, since in the US Census Bureau classification, the South contains a more demographically and politically diverse group of states than the North.

\section{Conclusion}
This study investigates how racial bias manifests itself in US legal opinions across regions and time periods. We observe implicit racial bias across almost all segments; commonly Black names are more likely to be associated with pre-classified ``unpleasant'' concepts, while commonly White names are more likely to be associated with pre-classified ``pleasant'' concepts. Opinions from Southern states do not show less change in racial bias across time than those from Northeastern states, and we do not find evidence that more recent legal opinions (post-1950) exhibit less implicit racial bias than older opinions (pre-1950). Our results suggest that aspects of the US judicial system may be unfair to Black individuals, warranting  further investigation. 

We offer several quantitative approaches for measuring implicit racial bias in a corpus and analyzing changes across corpora with differing data distributions. Our examination of changing name popularity demonstrates the need for research into less noisy signals for race in text. Future work could replicate these experiments with other approaches, as well as expand to other legal systems where the language of the law is important. Other related questions include whether bias changes around pivotal national events (e.g., 2020 George Floyd protests), or whether there is a correlation between the race of the judge and the race of the defendant or plaintiff. The results could indicate whether judges consider social context, or if they are partial to those belonging to the same race as them.

Overall, rigorous analyses of racial bias are key first steps toward achieving racial equality. 

\section{Limitations} 
This study relies on a relatively small set of racially distinctive names to signal race in judicial opinion texts. There may exist instances where a White individual's name is incorrectly categorized as commonly Black, or vice-versa. There is also a trade-off between using a large number of names (offering a weaker signal but producing more matches) vs. a small number (offering a stronger signal but producing fewer matches). We use the same curated set of names as many other studies, but recognize this is an imperfect option.

Names can also change in popularity over time. A consequence is that names which appear multiple times in a more recent corpus may not appear at all in an older one. Future work could create a set of the most popular racially distinctive names from each time period, perhaps by examining church/baptism records from historically Black and White denominations (this approach would overrepresent churchgoing Christians). More work is needed for identifying stronger signals for race, or leveraging a combination of signals.

The examined corpora had different case numbers. These discrepancies can be attributed to population shifts as well as the gradual addition of new states to the US. A future experiment could take a random sample of cases from more populated regions and employ sentence augmentation to generate synthetic data for smaller-sized corpora. This would keep corpus sizes even.

This study only examines bias between Black and White individuals. The US is a multiracial nation, so comprehensive analyses should incorporate other racial groups that have traditionally faced discrimination and consider intersectionality.

Lastly, our compilation of pleasant and unpleasant terms only incorporates the AFINN polarity and not the sentiment score in the WEAT calculation. Although this choice is consistent with previous work, a single word can have several polarities depending on context (e.g., ``the movie was sick'' vs. ``I am sick''). Given the formal nature of legal language and the stable results of Tables \ref{table:common_time} and \ref{table:common_region}, this fluctuation is likely less of an issue in legal text but may still influence our results. 

\section{Ethical Considerations} 
We are aware of the challenges with identifying signals for race in a corpus where it is not explicitly mentioned. This study utilizes a set of racially distinctive names proposed by \citet{butler_homola_2017} and utilized by \citet{islam-etal-2020-lexicon}, \citet{legalbias} and others. Studies like \citet{embeddings_bias} have also leveraged racially distinctive names. Such sets are imperfect signals, but they can point to implicit bias in text (as empirically proven in past studies). We hope future work will propose other ways to identify implicit textual bias.

\section*{Acknowledgements}
We thank Christiane Fellbaum for her generous support and mentorship for this paper. We also acknowledge support from the Princeton SEAS Friedland Independent Work/Senior Thesis Fund. 

\bibliography{anthology,custom}

\begin{thebibliography}{21}
\expandafter\ifx\csname natexlab\endcsname\relax\def\natexlab#1{#1}\fi

\bibitem[{Airport(2020)}]{censusregions}
Tampa~International Airport. 2020.
\newblock \href {https://www.tampaairport.com/us-census-regions} {Us census
  regions}.

\bibitem[{Bertrand and Mullainathan(2004)}]{bertrand_mullainathan_2004}
Marianne Bertrand and Sendhil Mullainathan. 2004.
\newblock \href {https://doi.org/10.1257/0002828042002561} {Are emily and greg
  more employable than lakisha and jamal? a field experiment on labor market
  discrimination}.
\newblock \emph{American Economic Review}, 94(4):991--1013.

\bibitem[{Butler and Homola(2017)}]{butler_homola_2017}
Daniel~M. Butler and Jonathan Homola. 2017.
\newblock \href {https://doi.org/10.1017/pan.2016.15} {An empirical
  justification for the use of racially distinctive names to signal race in
  experiments}.
\newblock \emph{Political Analysis}, 25(1):122–130.

\bibitem[{Friedman et~al.(2019)Friedman, Schmer-Galunder, Chen, and
  Rye}]{friedman-etal-2019-relating}
Scott Friedman, Sonja Schmer-Galunder, Anthony Chen, and Jeffrey Rye. 2019.
\newblock \href {https://doi.org/10.18653/v1/W19-3803} {Relating word embedding
  gender biases to gender gaps: A cross-cultural analysis}.
\newblock In \emph{Proceedings of the First Workshop on Gender Bias in Natural
  Language Processing}, pages 18--24, Florence, Italy. Association for
  Computational Linguistics.

\bibitem[{Greenwald et~al.(1998)Greenwald, McGhee, and
  Schwartz}]{Greenwald1998MeasuringID}
Anthony~G Greenwald, Debbie~E. McGhee, and Jordan L.~K. Schwartz. 1998.
\newblock Measuring individual differences in implicit cognition: the implicit
  association test.
\newblock \emph{Journal of personality and social psychology}, 74 6:1464--80.

\bibitem[{Hansford and Spriggs(2006)}]{hansford_spriggs_2006}
Thomas~G. Hansford and James~F. Spriggs. 2006.
\newblock \emph{The politics of precedent on the U.S. Supreme Court}.
\newblock Princeton University Press.

\bibitem[{Horowitz et~al.(2019)Horowitz, Brown, and Cox}]{pew}
Juliana~Menasce Horowitz, Anna Brown, and Kiana Cox. 2019.
\newblock \href
  {https://www.pewresearch.org/social-trends/2019/04/09/race-in-america-2019/}
  {Race in america 2019}.
\newblock Technical report, Pew Research Center.

\bibitem[{Islam et~al.(2016)Islam, Bryson, and Narayanan}]{embeddings_bias}
Aylin~Caliskan Islam, Joanna~J. Bryson, and Arvind Narayanan. 2016.
\newblock \href {http://arxiv.org/abs/1608.07187} {Semantics derived
  automatically from language corpora necessarily contain human biases}.
\newblock \emph{CoRR}, abs/1608.07187.

\bibitem[{Islam et~al.(2020)Islam, Xiao, and Mercer}]{islam-etal-2020-lexicon}
Jumayel Islam, Lu~Xiao, and Robert~E. Mercer. 2020.
\newblock \href {https://aclanthology.org/2020.lrec-1.380} {A lexicon-based
  approach for detecting hedges in informal text}.
\newblock In \emph{Proceedings of the 12th Language Resources and Evaluation
  Conference}, pages 3109--3113, Marseille, France. European Language Resources
  Association.

\bibitem[{Lauscher and Glava{\v{s}}(2019)}]{lauscher-glavas-2019-consistently}
Anne Lauscher and Goran Glava{\v{s}}. 2019.
\newblock \href {https://doi.org/10.18653/v1/S19-1010} {Are we consistently
  biased? multidimensional analysis of biases in distributional word vectors}.
\newblock In \emph{Proceedings of the Eighth Joint Conference on Lexical and
  Computational Semantics (*{SEM} 2019)}, pages 85--91, Minneapolis, Minnesota.
  Association for Computational Linguistics.

\bibitem[{Lax and Rader(2010)}]{Lax2010LegalCO}
Jeffrey~R. Lax and Kelly~T. Rader. 2010.
\newblock Legal constraints on supreme court decision making: Do
  jurisprudential regimes exist?
\newblock \emph{The Journal of Politics}, 72:273 -- 284.

\bibitem[{Levinson et~al.(2009)Levinson, Cai, and Young}]{levinson}
Justin Levinson, Huajian Cai, and Danielle Young. 2009.
\newblock Guilty by implicit racial bias: The guilty/not guilty implicit
  association test.
\newblock \emph{Ohio State Journal of Criminal Law}, 8.

\bibitem[{Merullo et~al.(2019)Merullo, Yeh, Handler, Grissom~II, O{'}Connor,
  and Iyyer}]{merullo-etal-2019-investigating}
Jack Merullo, Luke Yeh, Abram Handler, Alvin Grissom~II, Brendan O{'}Connor,
  and Mohit Iyyer. 2019.
\newblock \href {https://doi.org/10.18653/v1/D19-1666} {Investigating sports
  commentator bias within a large corpus of {A}merican football broadcasts}.
\newblock In \emph{Proceedings of the 2019 Conference on Empirical Methods in
  Natural Language Processing and the 9th International Joint Conference on
  Natural Language Processing (EMNLP-IJCNLP)}, pages 6355--6361, Hong Kong,
  China. Association for Computational Linguistics.

\bibitem[{Michel et~al.(2011)Michel, Shen, Aiden, Veres, Gray, Pickett,
  Hoiberg, Clancy, Norvig, Orwant, Pinker, Nowak, and Aiden}]{google_books}
Jean-Baptiste Michel, Yuan~Kui Shen, Aviva~Presser Aiden, Adrian Veres,
  Matthew~K. Gray, Joseph~P. Pickett, Dale Hoiberg, Dan Clancy, Peter Norvig,
  Jon Orwant, Steven Pinker, Martin~A. Nowak, and Erez~Lieberman Aiden. 2011.
\newblock \href {https://doi.org/10.1126/science.1199644} {Quantitative
  analysis of culture using millions of digitized books}.
\newblock \emph{Science}, 331(6014):176--182.

\bibitem[{Nielsen(2011)}]{AFINN}
Finn~{\AA}rup Nielsen. 2011.
\newblock A new anew: Evaluation of a word list for sentiment analysis in
  microblogs.
\newblock In \emph{\#MSM}.

\bibitem[{Pennington et~al.(2014)Pennington, Socher, and
  Manning}]{pennington-etal-2014-glove}
Jeffrey Pennington, Richard Socher, and Christopher Manning. 2014.
\newblock \href {https://doi.org/10.3115/v1/D14-1162} {{G}lo{V}e: Global
  vectors for word representation}.
\newblock In \emph{Proceedings of the 2014 Conference on Empirical Methods in
  Natural Language Processing ({EMNLP})}, pages 1532--1543, Doha, Qatar.
  Association for Computational Linguistics.

\bibitem[{Rice et~al.(2019)Rice, Rhodes, and Nteta}]{legalbias}
Douglas Rice, Jesse~H. Rhodes, and Tatishe Nteta. 2019.
\newblock \href {https://doi.org/10.1177/2053168019848930} {Racial bias in
  legal language}.
\newblock \emph{Research \& Politics}, 6(2):2053168019848930.

\bibitem[{Rice et~al.(2020)Rice, Rhodes, and Nteta}]{congressbias}
Douglas Rice, Jesse~H. Rhodes, and Tatishe Nteta. 2020.
\newblock Hidden in plain sight? implicit racial bias in the united states
  congress, 1967-2011.
\newblock (unpublished).

\bibitem[{Rice and Zorn(2021)}]{rice_zorn_2021}
Douglas~R. Rice and Christopher Zorn. 2021.
\newblock \href {https://doi.org/10.1017/psrm.2019.10} {Corpus-based
  dictionaries for sentiment analysis of specialized vocabularies}.
\newblock \emph{Political Science Research and Methods}, 9(1):20–35.

\bibitem[{Tiller and Cross(2005)}]{legaldoctrine}
Emerson Tiller and Frank Cross. 2005.
\newblock \href {https://doi.org/10.2139/ssrn.730284} {What is legal doctrine?}
\newblock \emph{Northwestern University Law Review}, 100.

\bibitem[{University(2018)}]{caselaw}
Harvard University. 2018.
\newblock \href {https://case.law} {Caselaw access project}.

\end{thebibliography}
\bibliographystyle{acl_natbib}

% \pdfoutput=1

% \documentclass[11pt]{article}
% \usepackage{needspace}
% % Remove the "review" option to generate the final version.
% \usepackage{acl}

% \usepackage{subcaption}

% % Standard package includes
% \usepackage{times}
% \usepackage{latexsym}
% \usepackage{float}
% \usepackage{amsmath}
% \usepackage{booktabs}
% % For proper rendering and hyphenation of words containing Latin characters (including in bib files)
% \usepackage[T1]{fontenc}

% % This assumes your files are encoded as UTF8
% \usepackage[utf8]{inputenc}

% % This is not strictly necessary, and may be commented out,
% % but it will improve the layout of the manuscript,
% % and will typically save some space.
% \usepackage{microtype}

% % Images/Graphics
% \usepackage{graphicx}

% % Equations
% \usepackage{amssymb, amsmath}

% % Check and X Mark
% \usepackage{pifont}
% \newcommand{\cmark}{\ding{51}}%
% \newcommand{\xmark}{\ding{55}}%

% % Tables
% \usepackage{tabularx}

% \usepackage{placeins}

% % \title{Appendices}

% \begin{document}
% \maketitle

\appendix

\section{Preprocessing Steps}
We run a standard preprocessing pipeline on each corpus to generate more meaningful embeddings, utilizing the spaCy \texttt{en\_core\_web\_sm} model. We tokenize the corpus text, lowercase each token, and apply part-of-speech tagging. We then lemmatize each token and remove stop words and punctuation using spaCy's built-in default list. We output the cleaned corpus to a text file. 

\section{Names Used}
\begin{figure}[hbt]
\begin{itemize}
\item \textbf{Traditionally White Names}: Abigail, Allison, Amy, Anne, Bradley, Brett, Caitlin, Carly,  Carrie, Claire, Cody, Cole, Colin, Connor, Dustin, Dylan, Emily, Emma, Garrett, Geoffrey, Greg, Hannah, Heather, Holly, Hunter, Jack, Jake, Jay, Jenna, Jill, Katelyn, Katherine, Kathryn, Katie, Kristen, Logan, Luke, Madeline, Matthew, Maxwell, Molly, Sarah, Scott, Tanner, Todd, Wyatt

\item \textbf{Traditionally Black Names}: Aaliyah, Alexus, Darius, Darnell, DeAndre, DeShawn, Deja, Dominique, Ebony, Jada, Jamal, Jasmin, Jasmine, Jazmine, Jermaine, Keisha, Kiara, LaShawn, Latonya, Latoya, Precious, Rasheed, Raven, Terrance, Tremayne, Tyrone, Xavier

\end{itemize}
\caption{Set of names used in this project, originally from Butler and Homola (2017).}
\label{fig:names_used}
\end{figure}

\Needspace{30\baselineskip}
\section{Word Embedding Association Test Results, Randomly Generated Corpora}

\begin{table}[hbt]
  \centering
  \setlength{\tabcolsep}{1.3em} % for the horizontal padding
  %{\renewcommand{\arraystretch}{1.2}% for the vertical padding
  \caption{Word embedding association test results for randomly generated corpora}
  \begin{tabular}{ccc}
    \toprule
    \textbf{File} & \textbf{Test Statistic} & \textbf{Significant?} \\
    \midrule
    0 & 6.4966 & \cmark \\
    1 & 2.6731 & \cmark \\
    2 & 3.7976 & \cmark \\
    3 & 3.2848 & \cmark \\ 
    4 & 3.4882 & \cmark \\
    5 & 3.6082 & \cmark \\
    6 & 2.5455 & \cmark \\
    7 & 3.9049 & \cmark \\
    8 & 3.8534 & \cmark \\
    9 & 2.6082 & \cmark \\
    10 & 5.0419 & \cmark \\
    11 & 4.3451 & \cmark \\
    12 & 3.3619 & \cmark \\
    13 & -1.4442 & \xmark \\
    14 & 4.1554 & \cmark \\
    15 & 3.7771 & \cmark \\
    16 & 3.8142 & \cmark \\
    17 & 1.5416 & \xmark \\
    18 & 2.0845 & \cmark \\
    19 & 4.2547 & \cmark \\
    \bottomrule
  \end{tabular}
  \label{table:random_stats}
\end{table}

% \newpage
% \Needspace{\baselineskip}
\onecolumn
\section{Case Counts}
\label{appendix:case_counts}
\begin{table}[htb!]
  \centering
  \caption{Case counts. Federal opinions from 1980-2009 are capped at 700,000 due to size constraints.}
  %\setlength{\tabcolsep}{1.3em} % for the horizontal padding
  %\setlength{\extrarowheight}{0.4em}
  %{\renewcommand{\arraystretch}{1.2}% for the vertical padding
  \begin{tabular}{rrrrrr}
    \toprule
      & \textbf{Northeast} & \textbf{South} & \textbf{Midwest} & \textbf{West} & \textbf{Federal}\\
    \midrule 
    \textbf{1860 - 1889} & 111799 & 82416 & 88295 & 17560 & 28123\\
    \textbf{1890 - 1919} & 343960 & 190129 & 191487 & 63960 & 68428\\
    \textbf{1920 - 1949} & 273399 & 293747 & 155295 & 78777 & 174836\\
    \textbf{1950 - 1979} & 295453 & 455959 & 145750 & 111250 & 340776\\
    \textbf{1980 - 2009} & 639186 & 791503 & 340022 & 248325 & 700000\\
    \bottomrule
  \end{tabular}
  \label{table:casecounts}
\end{table}

\newpage
\section{Popularity of Selected Names over Time}

\begin{figure*}[!htb]
    \centering
    \begin{subfigure}[b]{\linewidth}
    \centering
    \includegraphics[width=\linewidth]{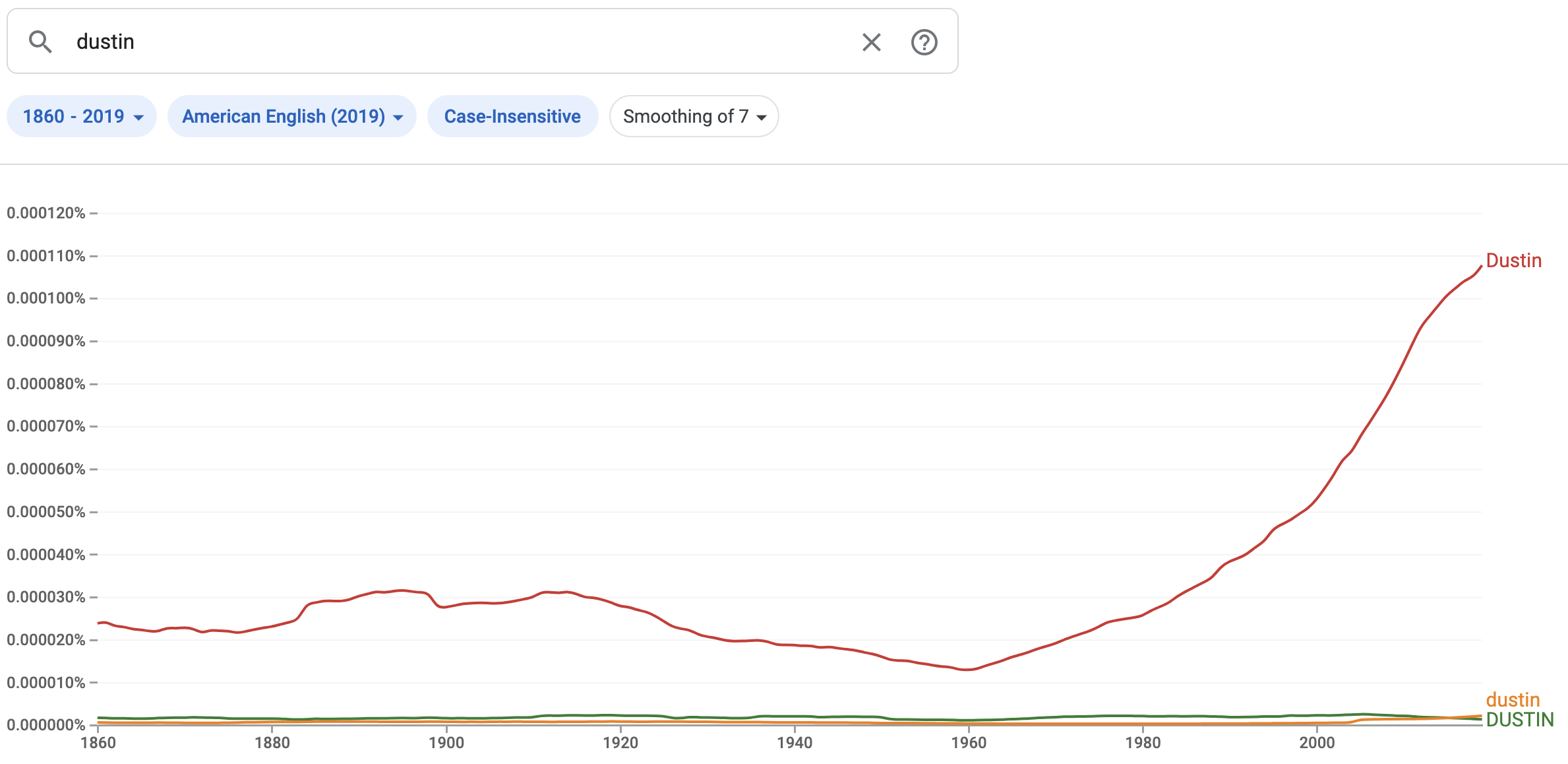}
    \caption{Recently popular White name: Dustin.}
    \label{fig:unpop_white}
    \end{subfigure}
    \hspace{0.1cm}
    \begin{subfigure}[b]{\linewidth}
    \centering
    \includegraphics[width=\linewidth]{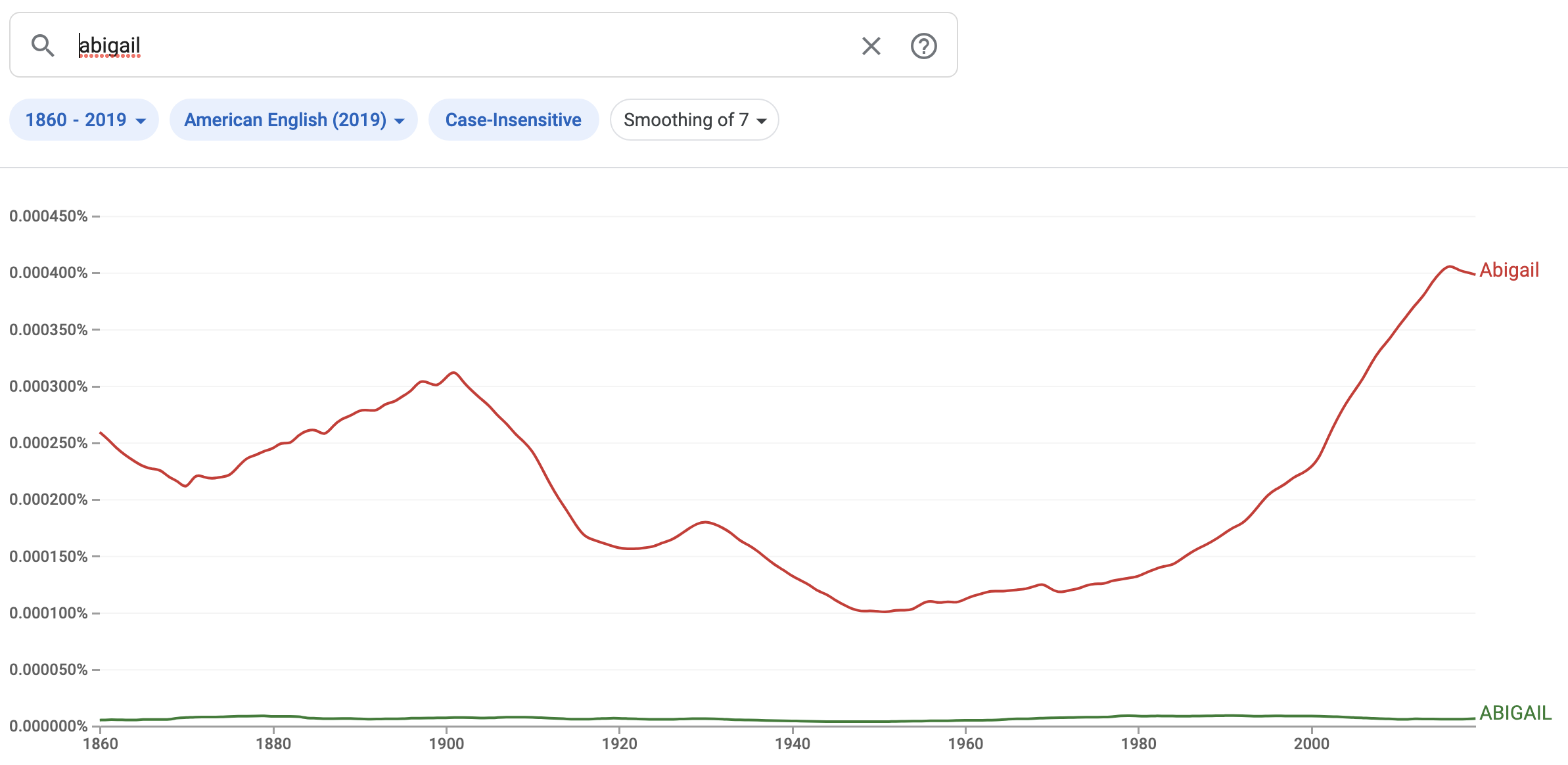}
    \caption{Consistently popular White name: Abigail.}
    \label{fig:pop_white}
    \end{subfigure}
    \caption{Examples of recently popular vs. consistently popular, traditionally White names.}
\end{figure*}
\begin{figure*}[hbt!]
    \centering
    \begin{subfigure}[b]{\linewidth}
    \centering
    \includegraphics[width=\linewidth]{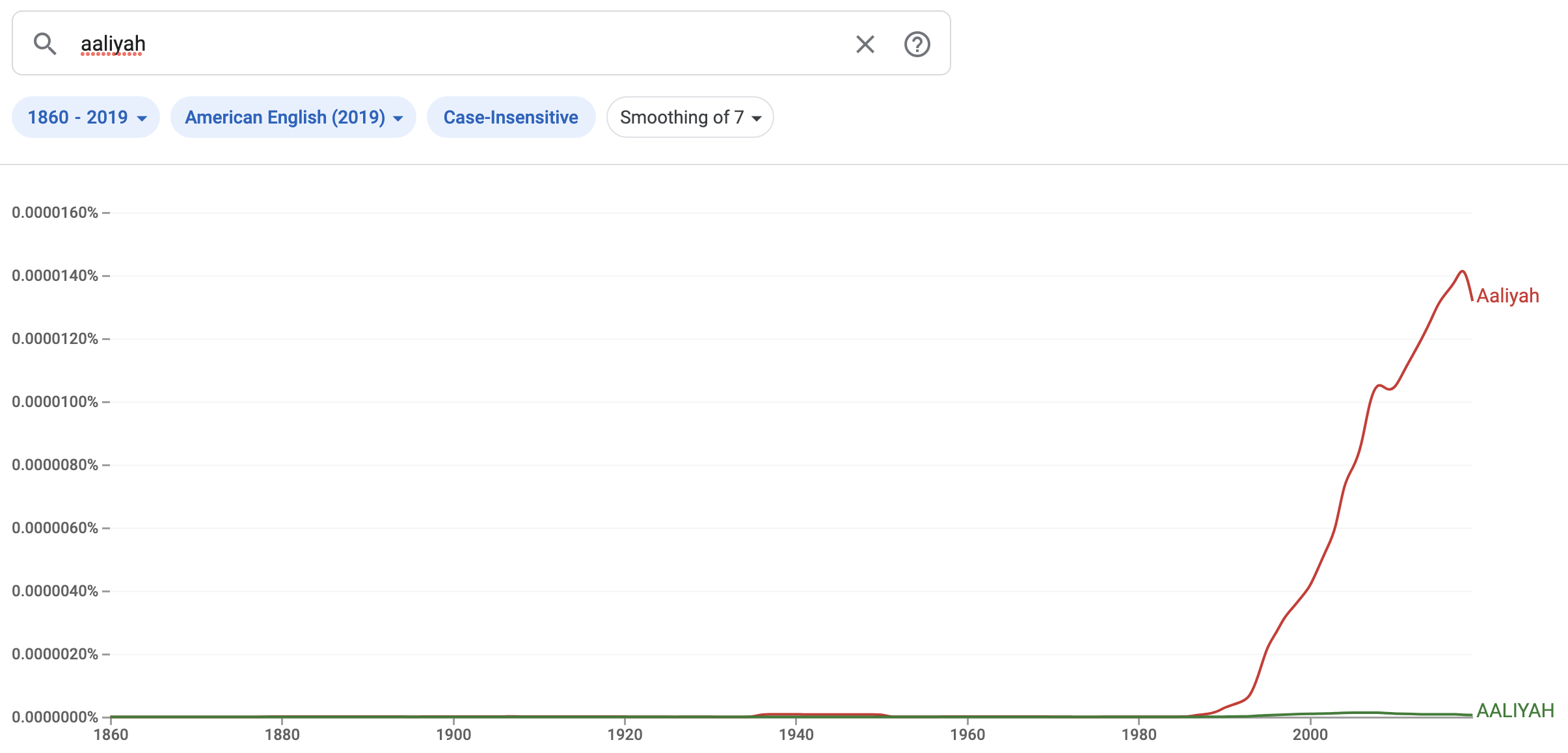}
    \caption{Recently popular, traditionally Black name: Aaliyah.}
    \label{fig:unpop_black}
    \end{subfigure}
    \hspace{0.1cm}
    \begin{subfigure}[b]{\linewidth}
    \centering
    \includegraphics[width=\linewidth]{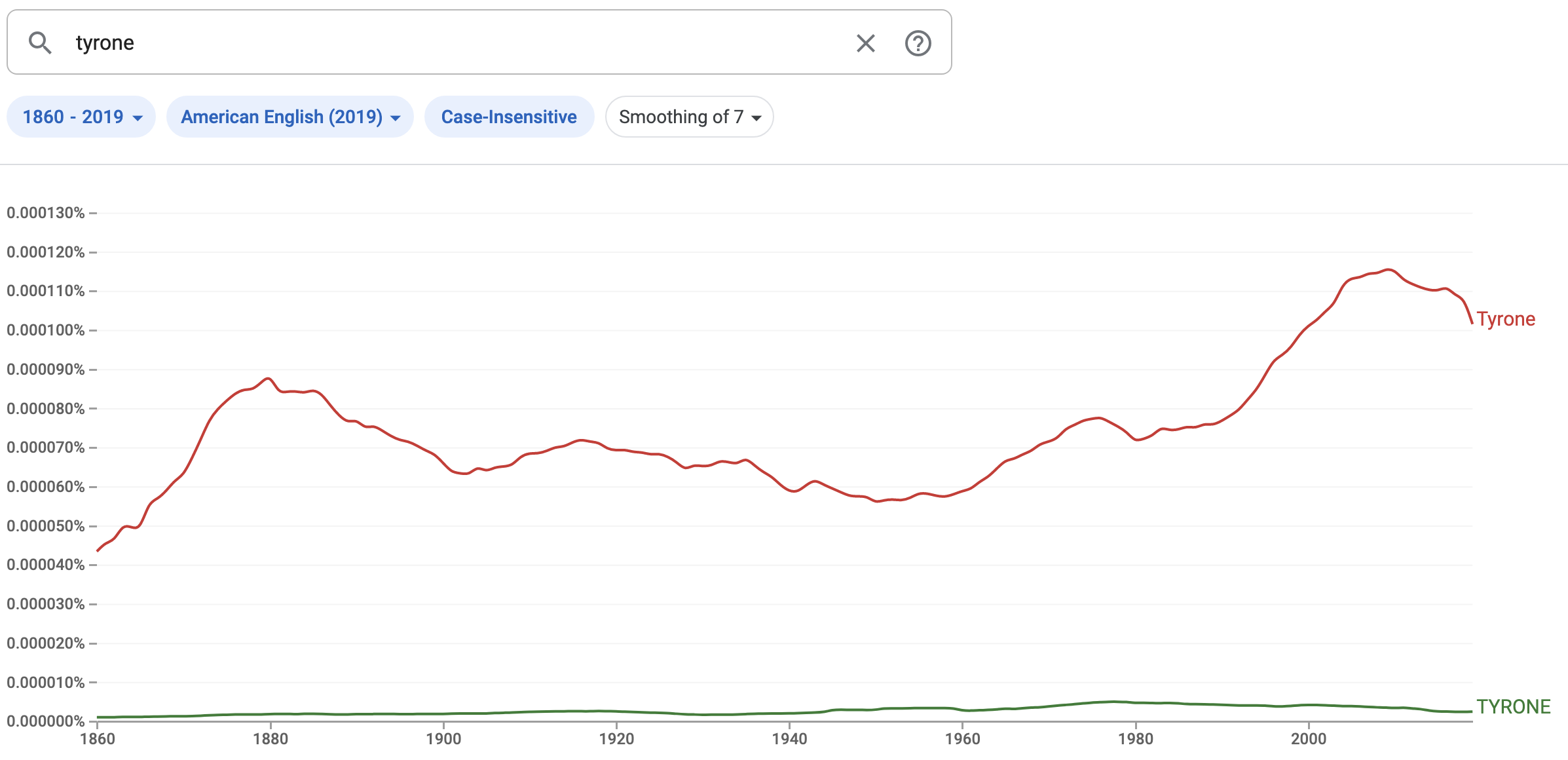}
    \caption{Consistently popular, traditionally Black name: Tyrone.}
    \label{fig:pop_black}
    \end{subfigure}
    \caption{Examples of recently popular vs. consistently popular, traditionally Black names.}
\end{figure*}

\newpage\phantom{blah}
\begin{figure*}[!hbt]
\section{Word Embedding Association Tests}
    \begin{minipage}{0.18\linewidth}
    \centering
    \ 1860-1889
    \end{minipage}
    \begin{minipage}{0.18\linewidth}
    \centering
    \ 1890-1919
    \end{minipage}
    \begin{minipage}{0.18\linewidth}
    \centering
    \ 1920-1949
    \end{minipage}
    \begin{minipage}{0.18\linewidth}
    \centering
    \  1950-1979
    \end{minipage}
    \begin{minipage}{0.18\linewidth}
    \centering
    \ 1980-2009
    \end{minipage}
    \begin{minipage}{0.18\linewidth}
    \centering
    \includegraphics[scale=0.22]{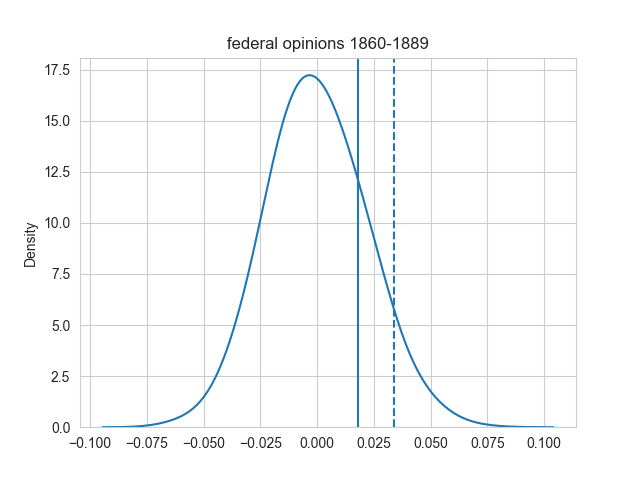}
    \vspace{-0.5cm}
    \caption*{ }
    \end{minipage}
    \hspace{0.13cm}
    \begin{minipage}{0.18\linewidth}
    \centering
    \includegraphics[scale=0.22]{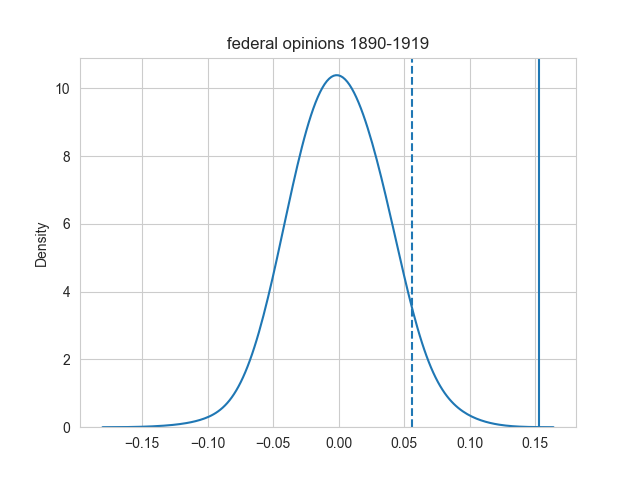}
    \vspace{-0.5cm}
    \caption*{ }
    \end{minipage}
    \hspace{0.13cm}
    \begin{minipage}{0.18\linewidth}
    \centering
    \includegraphics[scale=0.22]{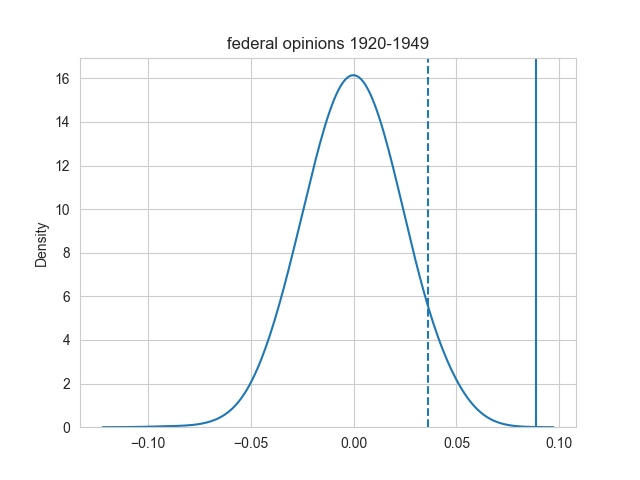}
    \vspace{-0.5cm}
    \caption*{\large Federal} % something here, description wise
    \end{minipage}
    \hspace{0.13cm}
    \begin{minipage}{0.18\linewidth}
    \centering
    \includegraphics[scale=0.22]{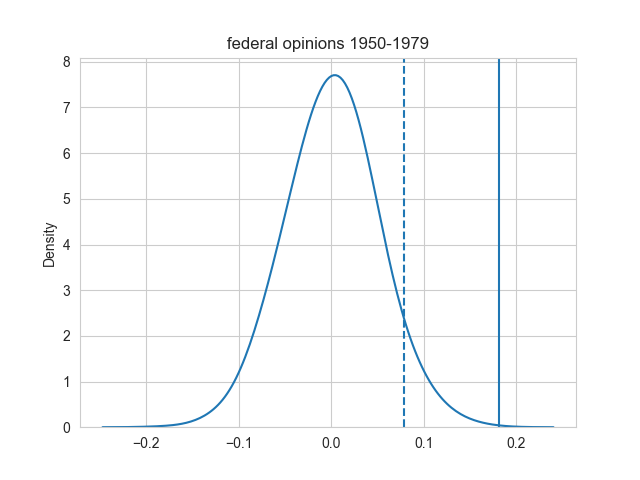}
    \vspace{-0.5cm}
    \caption*{ }
    \end{minipage}
    \hspace{0.13cm}
    \begin{minipage}{0.18\linewidth}
    \centering
    \includegraphics[scale=0.22]{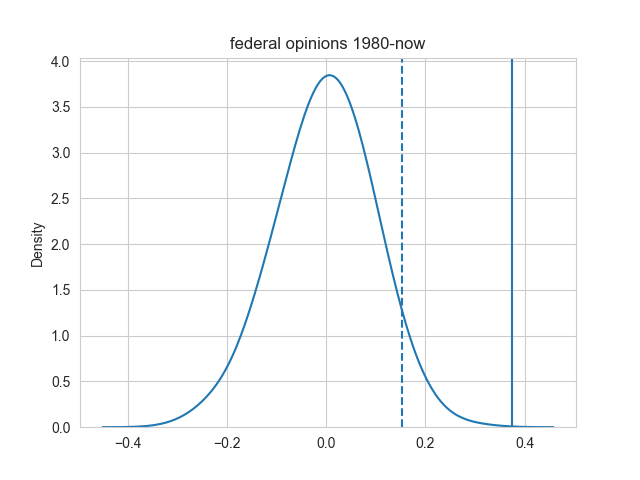}
    \vspace{-0.5cm}
    \caption*{ }
    \end{minipage}
    \vspace{0.1cm}

    \begin{minipage}{0.18\linewidth}
    \centering
    \includegraphics[scale=0.22]{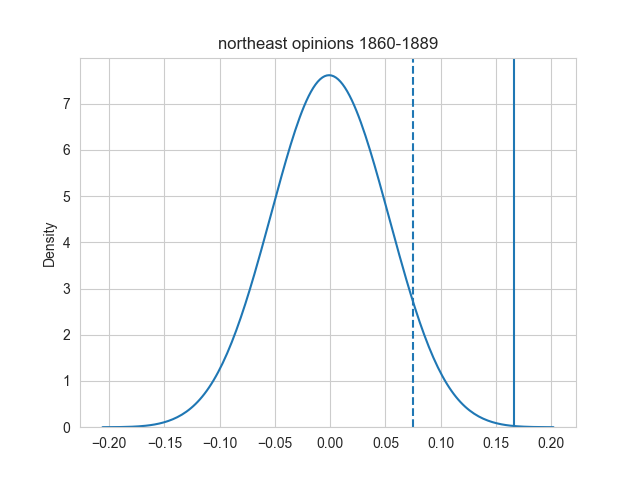}
    \vspace{-0.5cm}
    \caption*{ }
    \end{minipage}
    \hspace{0.13cm}
    \begin{minipage}{0.18\linewidth}
    \centering
    \includegraphics[scale=0.22]{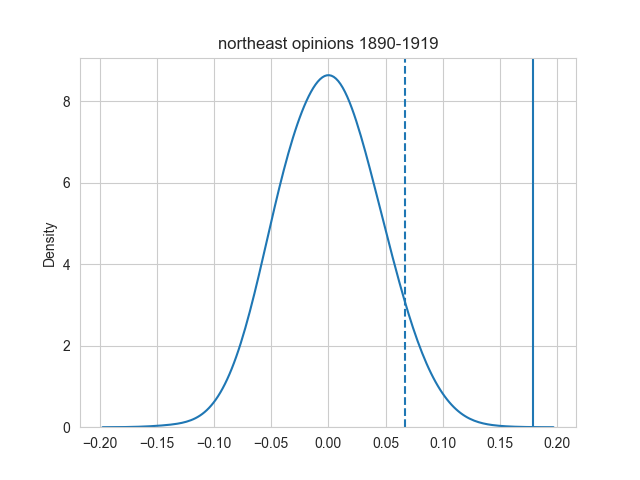}
    \vspace{-0.5cm}
    \caption*{ }
    \end{minipage}
    \hspace{0.13cm}
    \begin{minipage}{0.18\linewidth}
    \centering
    \includegraphics[scale=0.22]{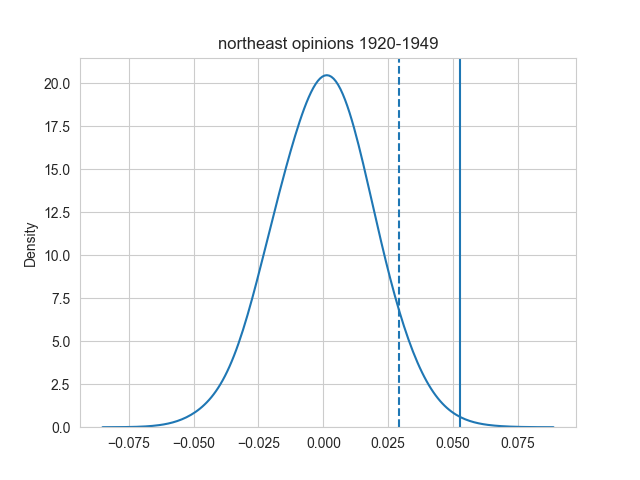}
    \vspace{-0.5cm}
    \caption*{\large Northeast}
    \end{minipage}
    \hspace{0.13cm}
    \begin{minipage}{0.18\linewidth}
    \centering
    \includegraphics[scale=0.22]{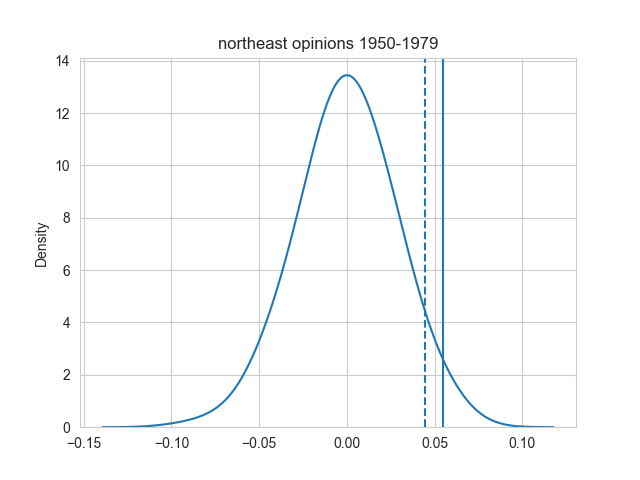}
    \vspace{-0.5cm}
    \caption*{ }
    \end{minipage}
    \hspace{0.13cm}
    \begin{minipage}{0.18\linewidth}
    \centering
    \includegraphics[scale=0.22]{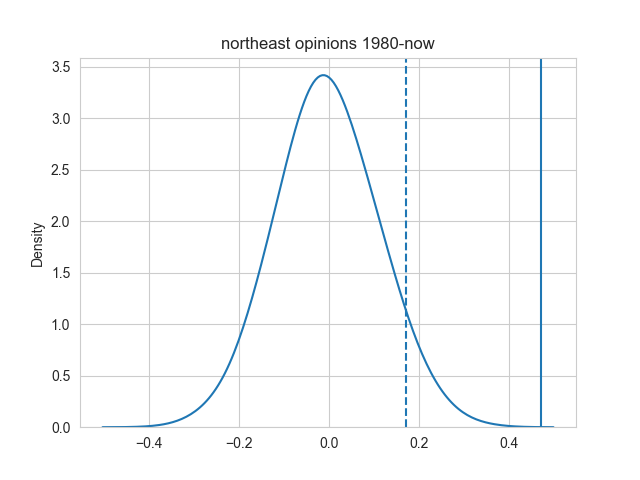}
    \vspace{-0.5cm}
    \caption*{ }
    \end{minipage}
    \vspace{0.1cm}
    
    \begin{minipage}{0.18\linewidth}
    \centering
    \includegraphics[scale=0.22]{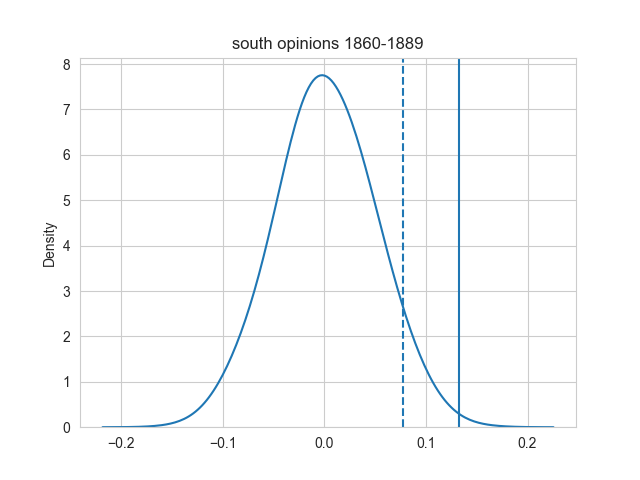}
    \vspace{-0.5cm}
    \caption*{ }
    \end{minipage}
    \hspace{0.13cm}
    \begin{minipage}{0.18\linewidth}
    \centering
    \includegraphics[scale=0.22]{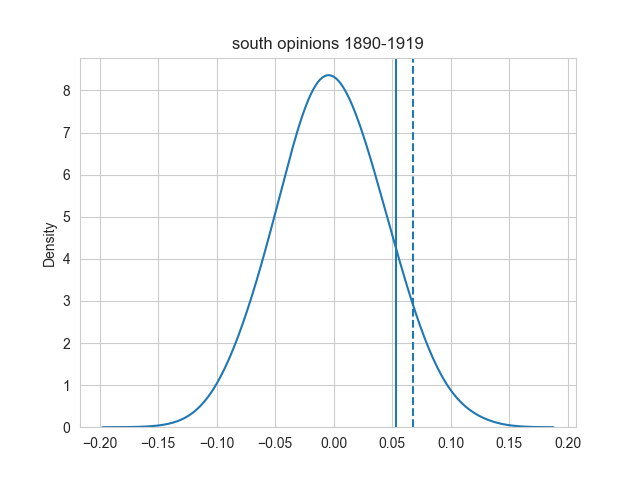}
    \vspace{-0.5cm}
    \caption*{ }
    \end{minipage}
    \hspace{0.13cm}
    \begin{minipage}{0.18\linewidth}
    \centering
    \includegraphics[scale=0.22]{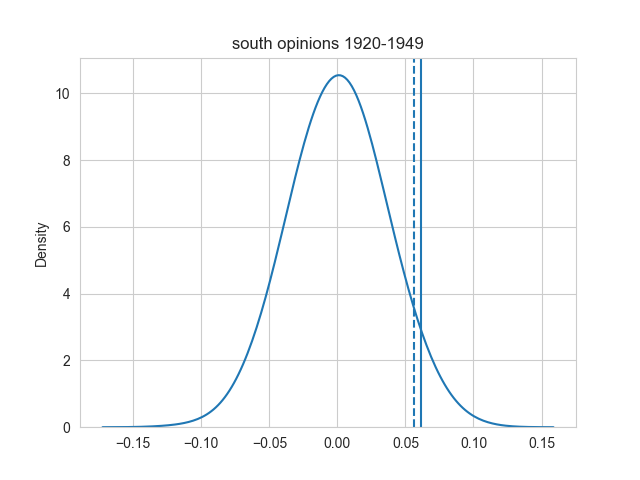}
    \vspace{-0.5cm}
    \caption*{\large South}
    \end{minipage}
    \hspace{0.13cm}
    \begin{minipage}{0.18\linewidth}
    \centering
    \includegraphics[scale=0.22]{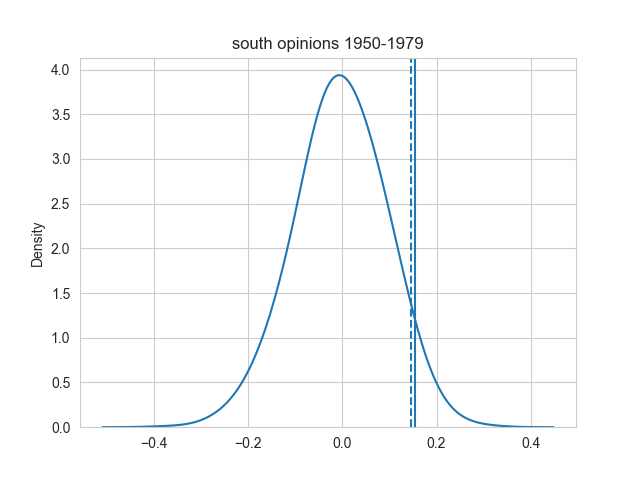}
    \vspace{-0.5cm}
    \caption*{ }
    \end{minipage}
    \hspace{0.13cm}
    \begin{minipage}{0.18\linewidth}
    \centering
    \includegraphics[scale=0.22]{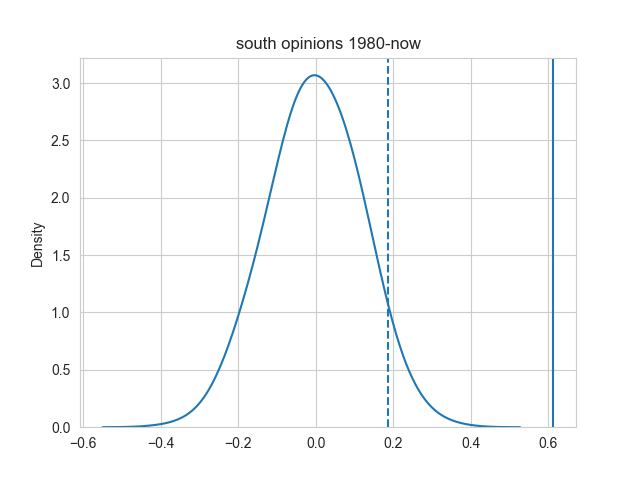}
    \vspace{-0.5cm}
    \caption*{ }
    \end{minipage}
    \vspace{0.1cm}
    
    \begin{minipage}{0.18\linewidth}
    \centering
    \includegraphics[scale=0.22]{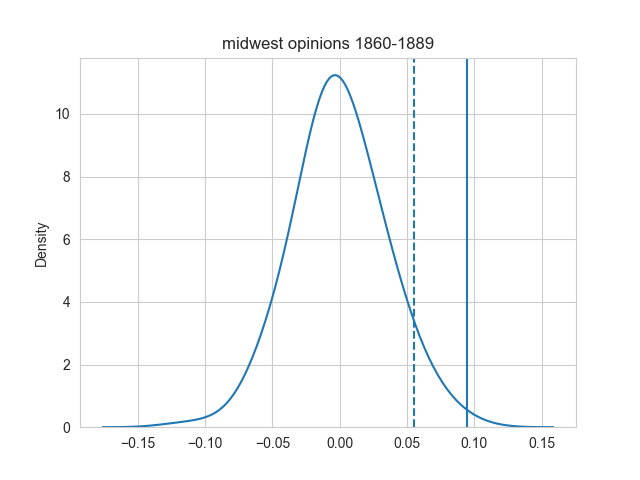}
    \vspace{-0.5cm}
    \caption*{ }
    \end{minipage}
    \hspace{0.13cm}
    \begin{minipage}{0.18\linewidth}
    \centering
    \includegraphics[scale=0.22]{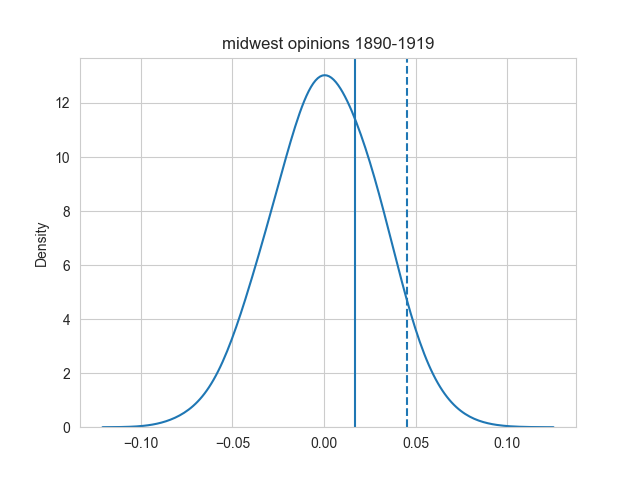}
    \vspace{-0.5cm}
    \caption*{ }
    \end{minipage}
    \hspace{0.13cm}
    \begin{minipage}{0.18\linewidth}
    \centering
    \includegraphics[scale=0.22]{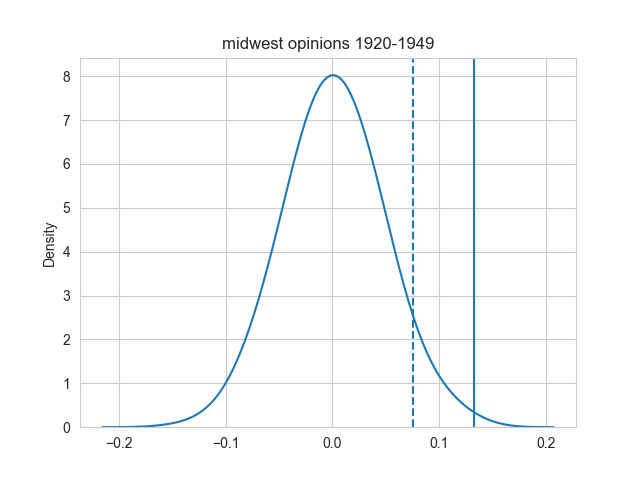}
    \vspace{-0.5cm}
    \caption*{\large Midwest}
    \end{minipage}
    \hspace{0.13cm}
    \begin{minipage}{0.18\linewidth}
    \centering
    \includegraphics[scale=0.22]{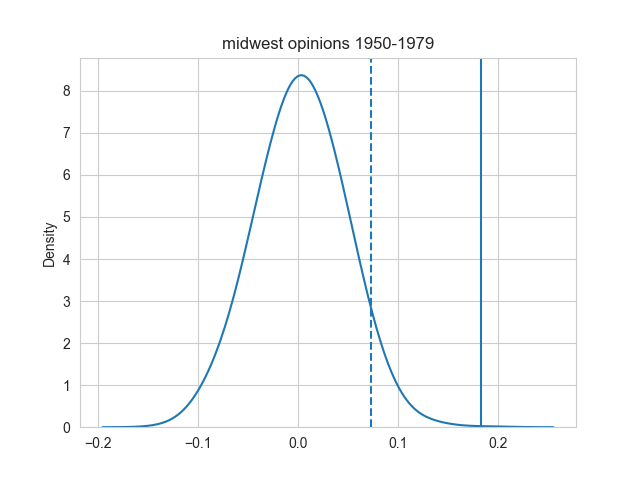}
    \vspace{-0.5cm}
    \caption*{ }
    \end{minipage}
    \hspace{0.13cm}
    \begin{minipage}{0.18\linewidth}
    \centering
    \includegraphics[scale=0.22]{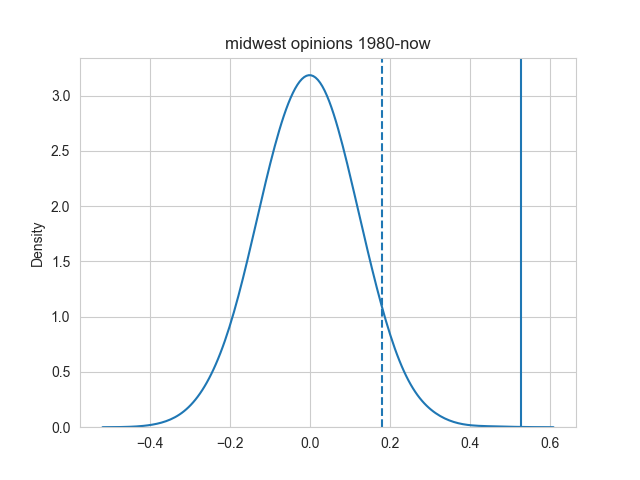}
    \vspace{-0.5cm}
    \caption*{ }
    \end{minipage}
    \vspace{0.1cm}
    
    \begin{minipage}{0.18\linewidth}
    \centering
    \includegraphics[scale=0.22]{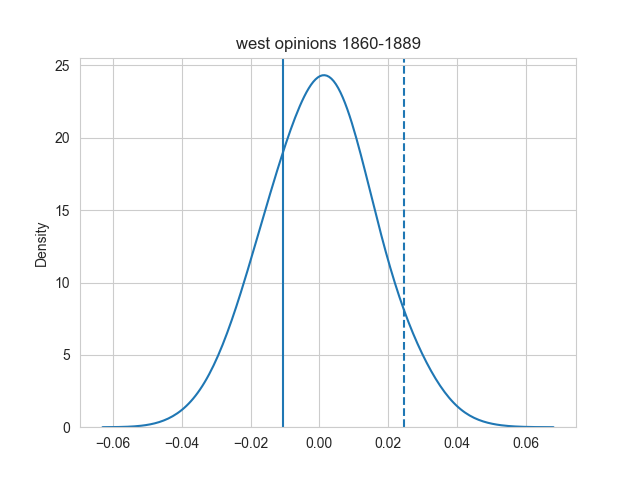}
    \vspace{-0.5cm}
    \caption*{ }
    \end{minipage}
    \hspace{0.13cm}
    \begin{minipage}{0.18\linewidth}
    \centering
    \includegraphics[scale=0.22]{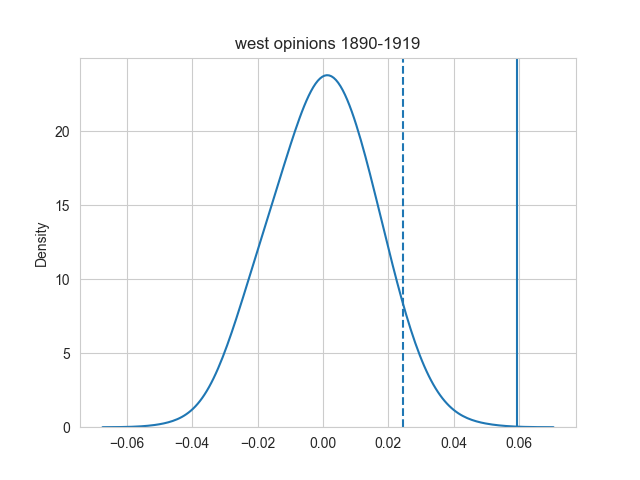}
    \vspace{-0.5cm}
    \caption*{ }
    \end{minipage}
    \hspace{0.13cm}
    \begin{minipage}{0.18\linewidth}
    \centering
    \includegraphics[scale=0.22]{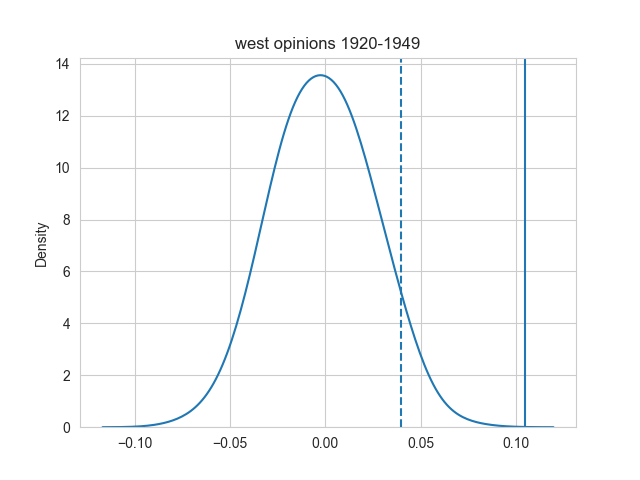}
    \vspace{-0.5cm}
    \caption*{\large West}
    \end{minipage}
    \hspace{0.13cm}
    \begin{minipage}{0.18\linewidth}
    \centering
    \includegraphics[scale=0.22]{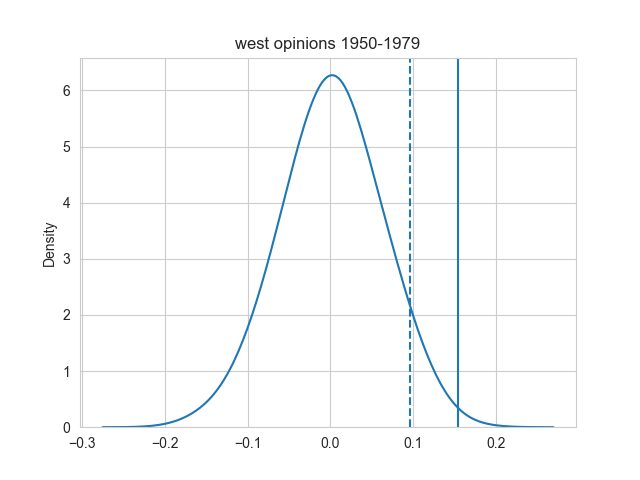}
    \vspace{-0.5cm}
    \caption*{ }
    \end{minipage}
    \hspace{0.13cm}
    \begin{minipage}{0.18\linewidth}
    \centering
    \includegraphics[scale=0.22]{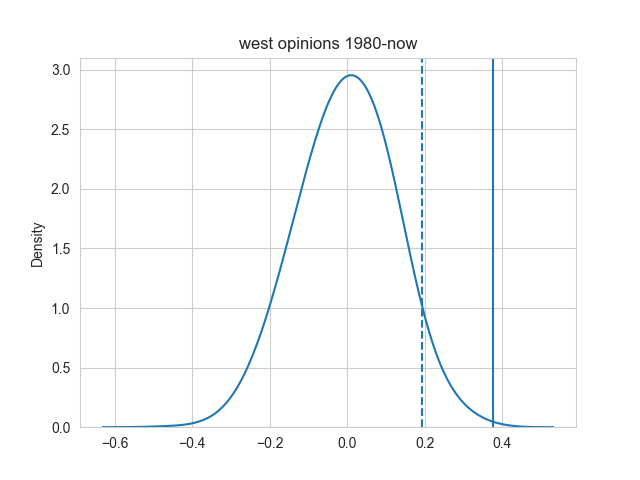}
    \label{fig:word_tests}
    \vspace{-0.5cm}
    \caption*{ }
    \end{minipage}
    \caption{Word embedding association tests by corpus.}
    \caption*{\centering The dotted line indicates the threshold for $\alpha = 0.05$, and the solid line indicates the test statistic for the original split of pleasant and unpleasant terms.}
\end{figure*}
\FloatBarrier

% \end{document}

\end{document}